\documentclass[conference]{IEEEtran}
\usepackage[letterpaper]{geometry}
\newgeometry{left=1.57cm,right=1.57cm,bottom=2.54cm,top=2.0cm} 

\IEEEoverridecommandlockouts
\usepackage{cite}
\usepackage{amsmath,amssymb,amsfonts,amsthm,enumitem}

\usepackage{graphicx}
\usepackage{textcomp}
\usepackage{float}
\usepackage{placeins}
\usepackage{multirow}
\usepackage{dblfloatfix}
\usepackage{pifont}
\usepackage[table,xcdraw]{xcolor}
\usepackage[normalem]{ulem}
\useunder{\uline}{\ul}{}
\usepackage{booktabs}
\usepackage[none]{hyphenat}

\usepackage[ruled,vlined]{algorithm2e}

\usepackage{adjustbox}
\usepackage{enumitem}
\usepackage[english]{babel}
\usepackage[utf8]{inputenc}
\usepackage[colorinlistoftodos]{todonotes}
\usepackage{balance}
\usepackage{comment}

\begin{document}
	
\sloppy	
\title{Exploring Fault-Energy Trade-offs in Approximate DNN Hardware Accelerators}
\author{\IEEEauthorblockN{Ayesha Siddique}
\IEEEauthorblockA{University of Missouri\\
Columbia, MO, USA\\
ayesha.siddique@mail.missouri.edu}
\and
\IEEEauthorblockN{Kanad Basu}
\IEEEauthorblockA{The University of Texas at Dallas\\
Dallas, TX, USA \\
kanad.basu@utdallas.edu}
\and
\IEEEauthorblockN{Khaza Anuarul Hoque}
\IEEEauthorblockA{University of Missouri\\
Columbia, MO, USA\\
hoquek@missouri.edu}
}
\maketitle
	\begin{abstract}

Systolic array-based deep neural network (DNN) accelerators have recently gained prominence for their low computational cost. However, their high energy consumption poses a bottleneck to their deployment in energy-constrained devices. To address this problem, approximate computing can be employed at the cost of some tolerable accuracy loss. However, such small accuracy variations may increase the sensitivity of DNNs towards undesired subtle disturbances, such as permanent faults. The impact of permanent faults in accurate DNNs has been thoroughly investigated in the literature.  Conversely, the impact of permanent faults in approximate DNN accelerators (AxDNNs) is yet under-explored. The impact of such faults may vary with the fault bit positions, activation functions and approximation errors in AxDNN layers. Such dynamacity poses a considerable challenge to exploring the trade-off between their energy efficiency and fault resilience in AxDNNs. Towards this, we present an extensive layer-wise and bit-wise fault resilience and energy analysis of different AxDNNs, using the state-of-the-art Evoapprox8b signed multipliers. In particular, we vary the stuck-at-0, stuck-at-1 fault-bit positions, and activation functions to study their impact using the most widely used MNIST and Fashion-MNIST datasets. Our quantitative analysis shows that the permanent faults exacerbate the accuracy loss in AxDNNs when compared to the accurate DNN accelerators. For instance, a permanent fault in AxDNNs can lead up to 66\% accuracy loss, whereas the same faulty bit can lead to only 9\% accuracy loss in an accurate DNN accelerator. Our results demonstrate that the fault resilience in AxDNNs is orthogonal to the energy efficiency.

	\end{abstract}
	

	\maketitle 
\section{Introduction}
\label{sec:introduction}
Deep Neural Networks (DNNs) are widely popular in big-data analytic applications such as, smart healthcare and telematics. Their anticipated adoption facilitates non-erroneous data-intensive classification, segmentation and translation. However, high precision often compels DNNs to use huge parameters space \cite{nguyen2017iris} with massive matrix multiplications. Such computational intensiveness limits their deployment on energy-constrained devices. For example, heavy edge analytics in battery-driven self-driving cars, where safety and energy are critical considerations, can lead to their unexpected energy outage. For enhancing the energy efficiency of such devices, many energy-aware systolic array-based DNN accelerators have been recently developed \cite{venieris2018toolflows} but their large size requirement for fast data processing poses meager energy gains in energy-constrained devices \cite{imani2018rapidnn}. This problem can be addressed with approximate computing that trades the accuracy of an application-specific system, by exploiting its intrinsic error resilience, for energy savings \cite{siddique2018approxcs}. It incorporates loop skipping and sampling rate reduction at the software level, and adopts inexact arithmetic units (e.g., multipliers and adders with truncated carry chains \cite{hanif2019cann} or bit-wise structural modifications \cite{riaz2020caxcnn}) and memory skipping \cite{song2019approximate} at the hardware level. Since, these practices are error inducing in nature \cite{marchisio2020red}; they are deemed to be more sensitive towards undesired subtle disturbances such as, permanent (hard errors) and transient faults (soft errors) \cite{rodrigues2019assessing}. 

A fault, along with approximation error, in any region of systolic array-based approximate DNN accelerators \cite{hanif2019cann} (AxDNNs) may propagate unevenly to multiple DNN layers. For example, an un-masked fault in the most significant bit (MSB), of their approximate multiplier's output, can cause deviation from the expected output of all AxDNN layers due to reusable nature of systolic arrays. However, the impact of a fault may vary according to the activation functions, fault type and bit position, and approximation error resilience of each layer. Since, the permanent faults affect the performance of accurate DNNs more significantly than occasional transient faults \cite{kundu2020high}. Their impact may be more prominent in AxDNNs due to their inexact nature. Recently, Kundu and Zhang et al. explored the impact of permanent faults on multiple locations of the accurate DNNs \cite{kundu2020high} \cite{zhang2018analyzing}. Hong et al. elucidated their limits for different bit-wise parameter corruptions \cite{hong2019terminal}. Li et al. exploited the fault tolerance of the accurate DNNs on the basis of different data types and number of layers \cite{li2017understanding}.  In other words, the state-of-the-art research is focused on the fault resilience analysis of accurate DNN accelerators. \textit{However, an extensive fault resilience analysis of AxDNNs and its relationship with energy efficiency is yet under-explored}. 


Towards this, we present an extensive permanent fault resilience and energy analysis of AxDNNs. Since, the multipliers consume more energy than other arithmetic units (e.g., adders) in DNNs \cite{marchisio2020red}. Therefore, we use their approximate counterparts with faults injection in their output bits. In particular, we benchmark the impact of stuck-at-0 and stuck-at-1 faults on approximate feed-forward neural networks (FFNNs), using the state-of-the-art Evoapprox8b signed multipliers \cite{mrazek2017evoapproxsb}, by varying the fault-bit position (least significant bit `LSB' to MSB) and incorporating different activation functions (e.g., \textit{tanh}, \textit{sigmoid}) in their hidden layers to foster further progress in this field. We explore the layer-wise fault resilience of AxDNNs, using the MNIST \cite{deng2012mnist} and Fashion MNIST \cite{xiao2017fashion} datasets to demonstrate the contribution of each layer towards the accuracy loss in the presence of approximation errors. Our quantitative fault resilience analysis shows that the faults exacerbates the accuracy loss with approximate computing in AxDNNs. For instance, a permanent fault in AxDNNs can lead up to 66\% accuracy loss, whereas the same fault in the same position can lead to only 9\% accuracy loss in an accurate DNN accelerator. Our results demonstrate that the fault resilience is orthogonal to the energy efficiency in of AxDNNs and varies from output to input layer in accordance with the type of activation functions and the amount of approximation error. The higher the approximation error is, the higher is the faults tendency to disrupt the output quality. 


The remainder of this paper is structured as follows: Section \ref{sec:prelim} provides the preliminary information about the fault resilience and state-of-the-art approximate multipliers. Section \ref{sec:methodology} presents our evaluation methodology. Section \ref{sec:results} discusses the results of fault resilience, and energy analysis of AxDNNs. Finally, Section \ref{sec:conclusion} concludes the paper.

\section{Background}
\label{sec:prelim}
This section provides a brief overview of the fault resilience and state-of-the-art approximate multipliers for better understanding of the paper.




\subsection{Fault Resilience}
\label{subsec:scores}
Fault Resilience is an important characteristic of DNNs that can be defined as a function of accuracy loss. The DNN accelerators inherit such resilience for a considerable range of fault-bits to ensure reliable computation \cite{kundu2020high}. However, it may become insignificant, when faults incur performance degradation by influencing the MSBs, especially, such that they remain unmasked in the resulting outputs. In other words, the bits may be stuck-at-0 or stuck-at-1 due to some hardware defects; however, they propagate to the output DNN layer only when their presence changes the value of the corresponding byte. For example, the stuck-at-1 fault at first bit position in the 4-bit output `0001' of a multiplier may remain masked (no effect) but at fourth bit position, it may lead to difference of 8 digits (in decimal) i.e., `1001'. Such output deviations may propagate as faults through multiple DNN layers and affect the application layer to a larger extent \cite{zhang2018analyzing}. It is important to note that unlike approximation errors, the faults are unexpected changes in the bits. In this paper, the stuck-at-0 and stuck-at-1 faults are used for fault resilience analysis of DNN accelerators.

\subsection{Approximate Multipliers}
\label{subsubsec:approxmult}
Approximate multipliers relax the abstraction of near-perfect accuracy in digital applications \cite{javed2018approxct} by simplifying the partial product matrix \cite{el2017embracing}, incorporating approximate counters or compressors in the partial product tree \cite{hanif2019cann} and truncating the carry propagation chain in partial products generation \cite{ullah2018smapproxlib} for low latency and power or energy consumption. However, the later approach often leads to high (undesired) truncation error and hence, requires error correction units \cite{esposito2018approximate}. Recently, Mrazek et al. developed a library of approximate multipliers, known as `Evoapprox8b' \cite{mrazek2017evoapproxsb}, which poses resource efficiency with tolerable accuracy loss in the most error resilient applications. It uses multi-objective Cartesian genetic programming for generating a set of ASIC-oriented optimal approximate multiplication circuits \cite{ullah2018smapproxlib}. It also contains some optimized accurate multipliers. In this paper, we used both accurate and approximate Evoapprox8b \cite{mrazek2017evoapproxsb} signed multipliers for comparative fault resilience and energy analysis of accurate DNNs and AxDNNs. 



\section{Evaluation Methodology}
\label{sec:methodology}

\begin{figure}[!h]
	\centering
	\vspace{-0.1in}
	\includegraphics[width=1\linewidth]{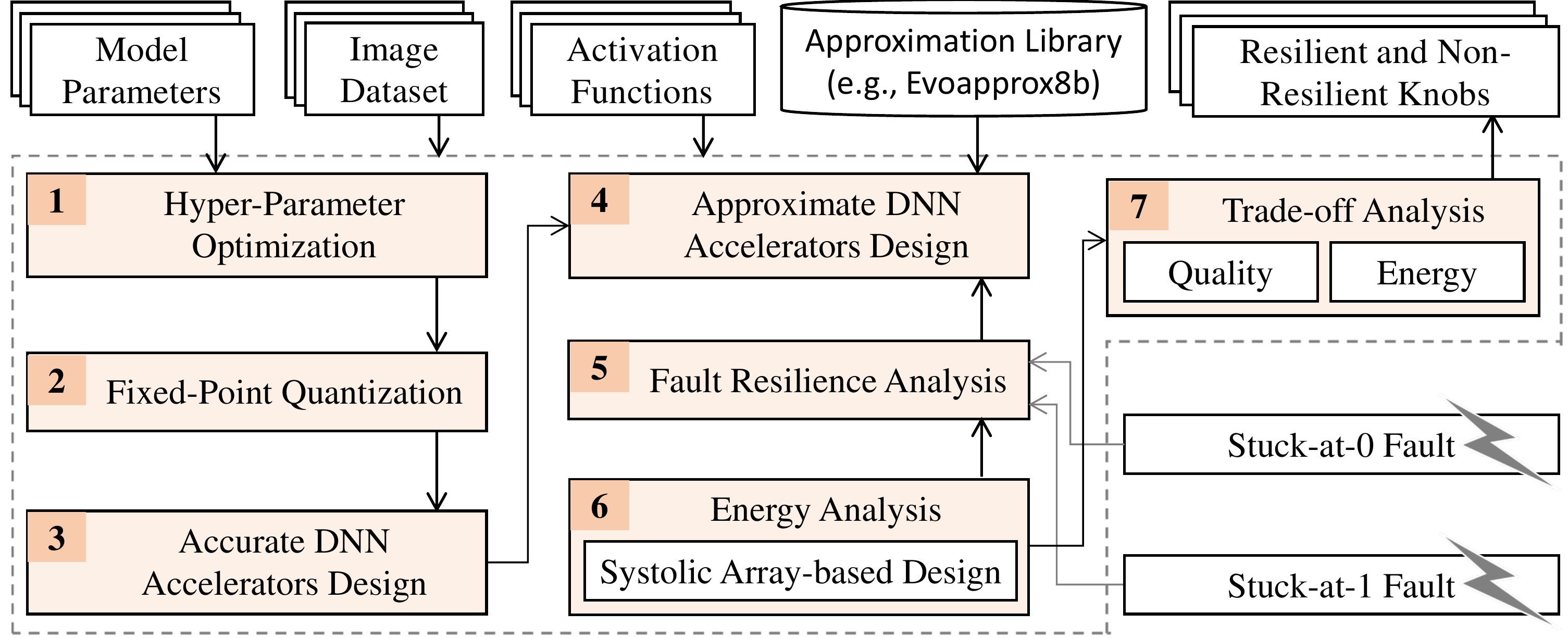}
	\caption{Evaluation Methodology for Analyzing Fault-Energy Trade-offs in Systolic Array-based DNN Architectures}
	\label{fig:methodlogy}
	\vspace{-0.25in}
\end{figure}

The performance of DNNs is mainly determined by their design parameters e.g., number of hidden layers, neurons, etc. Therefore, an optimal parameters configuration is first searched by trial and error method for the accurate DNN training and inference prior to 8-bit signed integer quantization. Then, the accurate multipliers are replaced with their approximate counterparts, using an open-source approximation library (e.g., Evoapprox8b \cite{mrazek2017evoapproxsb}), for approximation error resilience analysis. Next, the faults are injected in their different output bit positions (i.e, from LSB to MSB) and layers of AxDNNs. This fault injection is performed for different activation functions in the hidden layers to explore the dependency of the fault resilience on the activation functions. Lastly, the fault-energy trade-offs are analyzed by finding the energy consumption of AxDNNs, with systolic array-based hardware implementation, using a logic synthesis tool `Synopsys Design Compiler'. This trade-off analysis results in multiple energy-aware and, fault-resilient and non-resilient knobs. Fig. \ref{fig:methodlogy} presents an overview of our evaluation methodology.

\section{Results and Discussions}
\label{sec:results}
This section discusses the experimental setup and results for approximation error and fault resilience, and energy analysis of accurate DNNs and AxDNNs using the Evoapprox8b \cite{mrazek2017evoapproxsb} library. In this paper, the accurate multiplier refers to KV8 and approximate multipliers refers to KVA, KVB, KX2, KRC, KR6, L2H, L2D, L1G, KTY and KVL signed multipliers obtained from the Evoapprox8b library.

\subsection{Experimental Setup}
\subsubsection{Datasets}
In this paper, the fault resilience of accurate DNNs and AxDNNs is analyzed using two widely used datasets namely, the MNIST \cite{deng2012mnist} and Fashion-MNIST \cite{xiao2017fashion} datasets. These datasets contain 60,000 training and 10,000 test images. The MNIST dataset represents the labelled handwritten digits ranging from 0 to 9, whereas the Fashion MNIST is associated with labels from 10 different classes such as boots, shirts, etc.


\subsubsection{Model Configurations}
In this paper, the DNNs with design configurations 784-256-256-256-10 and 784-512-512-512-10 are used for the MNIST \cite{deng2012mnist} and Fashion MNIST \cite{xiao2017fashion} classification, respectively. For each of these configurations, two architectures with different activation functions are evaluated, denoted as Arch. 1 and Arch. 2. The input and hidden layers of Arch. 1 and Arch. 2 incorporate \textit{tanh} and \textit{sigmoid} as activation functions, respectively. Their output layer contains softmax as the activation function.

\begin{figure}[!b]
	\centering
	\vspace{-0.15in}
	\includegraphics[width=1\linewidth]{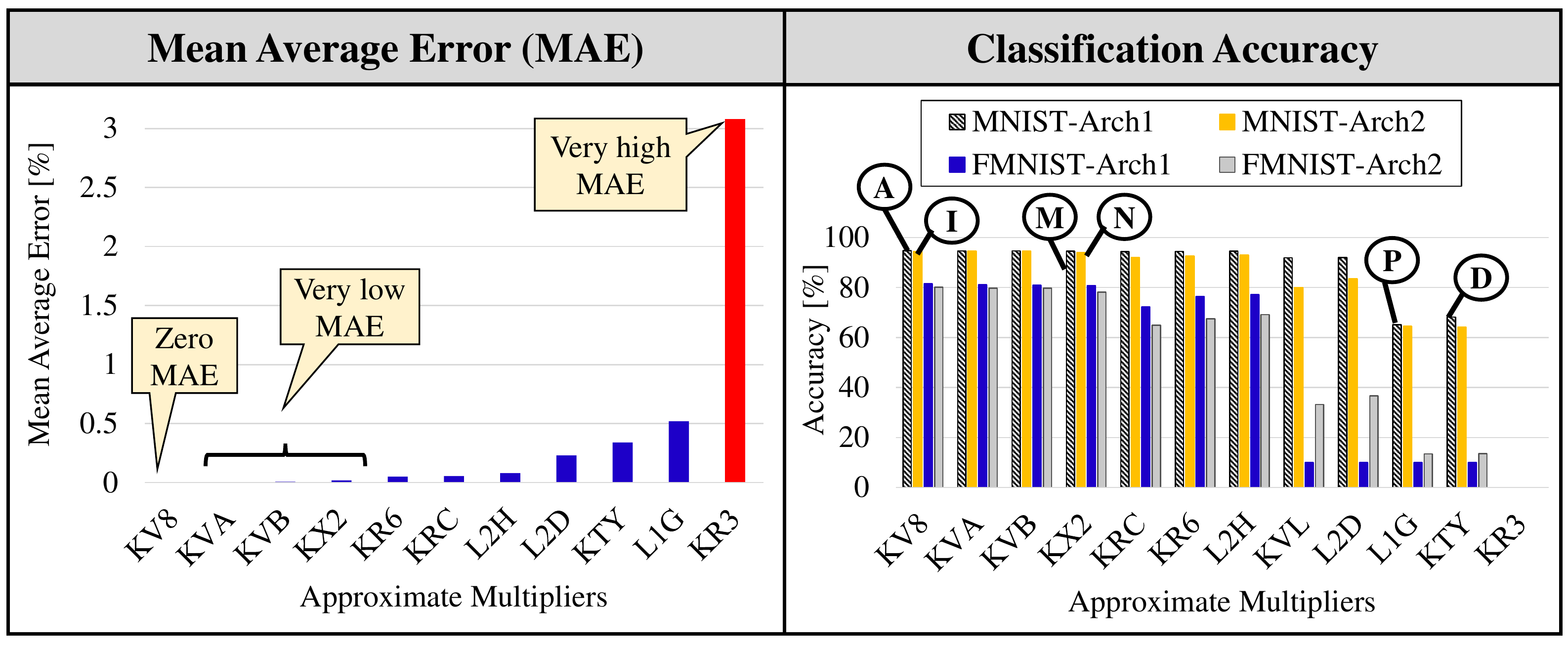}
	\caption{(a) Mean Average Error (MAE) of 8-bit Evoapprox8b \cite{mrazek2017evoapproxsb} Signed Multipliers, (b) Classification Accuracy of Approximate DNN architectures (AxDNN), using the MNIST \cite{deng2012mnist} and Fashion MNIST \cite{xiao2017fashion} datasets}
	\label{fig:maeAndAcc}. 
	\vspace{-0.1in}
\end{figure}

\subsection{Fault Resilience Analysis}
\label{faultRes}
In this section, the fault resilience of accurate DNN and AxDNN architectures is extensively studied with different fault configurations (fault types and bit positions).

\subsubsection{Approximation Error vs. Fault Resilience} 
\begin{figure*}[!b]
	\centering
	\vspace{-0.2in}
	\includegraphics[width=1\linewidth]{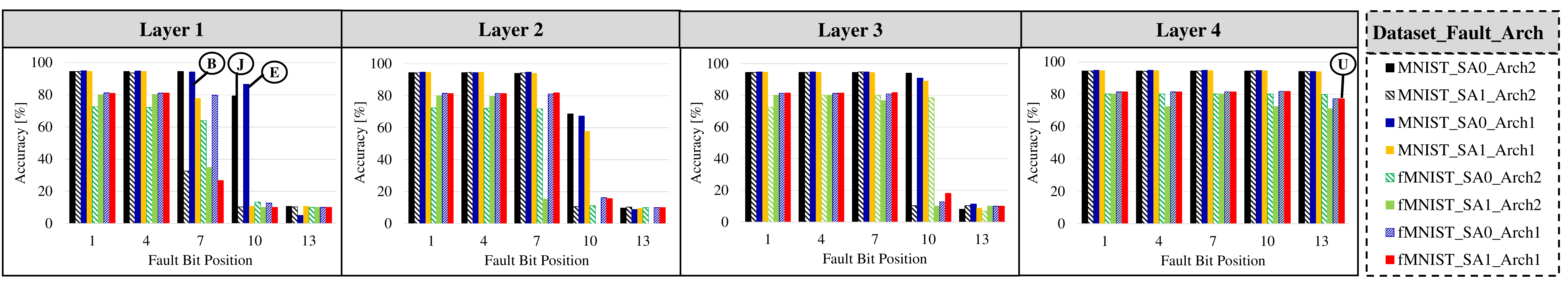}
	\caption{Stuck-at-0 and Stuck-at-1 Faults Resilience Analysis of Accurate DNN architectures using the MNIST \cite{deng2012mnist} and Fashion MNIST \cite{xiao2017fashion} datasets.}
	\label{fig:KV8}
	\vspace{-0.1in}
\end{figure*}

\begin{figure*}[!b]
	\centering
	\vspace{-0.01in}
	\includegraphics[width=1\linewidth]{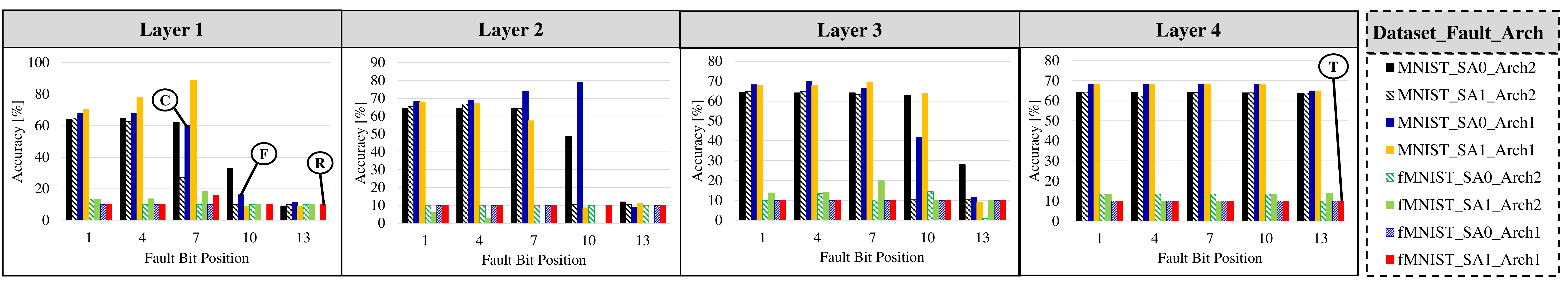}
	\caption{Stuck-at Faults Resilience Analysis of KTY-based Approximate DNNs (AxDNN) \cite{mrazek2017evoapproxsb}, using the MNIST \cite{deng2012mnist} and Fashion MNIST \cite{xiao2017fashion} datasets.}
	\label{fig:KTY}
	\vspace{-0.1in}
\end{figure*}

\begin{figure*}[!b]
	\centering
	\vspace{-0.25in}
	\includegraphics[width=1\linewidth]{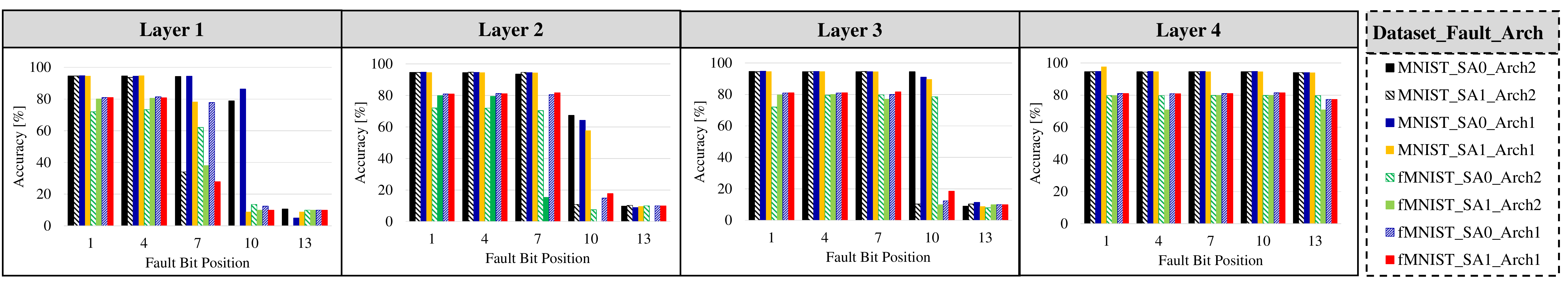}
	\caption{Stuck-at Faults Resilience Analysis of KVB-based Approximate DNNs (AxDNN) \cite{mrazek2017evoapproxsb}, using the MNIST \cite{deng2012mnist} and Fashion MNIST \cite{xiao2017fashion} datasets.}
	\label{fig:KVB}
	\vspace{-0.1in}
\end{figure*}

\begin{figure*}[!b]
	\centering
	\vspace{-0.01in}
	\includegraphics[width=1\linewidth]{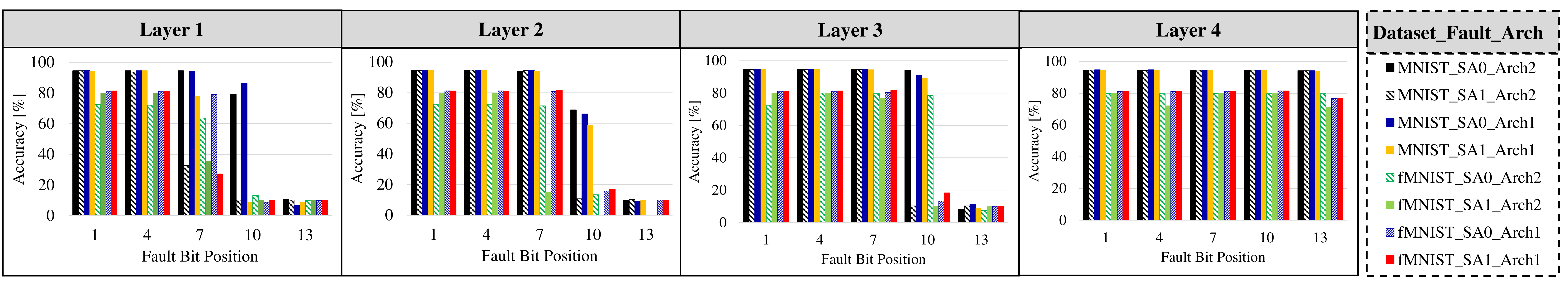}
	\caption{Stuck-at Faults Resilience Analysis of KVA-based Approximate DNNs (AxDNN) \cite{mrazek2017evoapproxsb}, using the MNIST \cite{deng2012mnist} and Fashion MNIST \cite{xiao2017fashion} datasets.}
	\label{fig:KVA_v2}
	\vspace{-0.1in}
\end{figure*}

In Fig. \ref{fig:maeAndAcc}, the mean average error (MAE) comparison of different multipliers shows that the L2D, KTY, L1G and KR3 approximate multipliers have high approximation error and hence, pose low classification accuracy in AxDNNs (see Fig. \ref{fig:maeAndAcc} (b)). The faults along-with such inexact nature of AxDNNs may render an application to perform below its error tolerance range. For example, the MNIST \cite{deng2012mnist} classification with stuck-at-0 fault in the seventh bit of the accurate multiplier in Arch 1. (fault in layer 1 only) results in 0.62\% accuracy loss in comparison to the corresponding non-faulty accurate DNN (see label A in Fig. \ref{fig:maeAndAcc} (b) and label B in Fig. \ref{fig:KV8}). However, the same fault configuration causes 34.49\% accuracy loss in the KTY (having MAE 0.34\% that is higher than accurate multiplier) based AxDNN (see label A in Fig. \ref{fig:maeAndAcc} (b) and C in  Fig. \ref{fig:KTY}). Here, the faults contribute towards approximately 8\% accuracy loss because the accuracy drops from 68.19\% to 60.32\% with fault injection (see label C in Fig. \ref{fig:KTY} and label D in Fig. \ref{fig:maeAndAcc} (b)). The same fault trend is observed in case of Fashion-MNIST classification. Our quantitative fault resilience analysis in Fig. \ref{fig:KVB} and Fig. \ref{fig:KVA_v2} show that the KVA- and KVB-based AxDNNs behave quite similar to accurate DNNs in the absence (see Fig. \ref{fig:KV8}) and presence of faults (see Fig. \ref{fig:maeAndAcc} (b)) due to their very low approximation error e.g., 0.0018\% and 0.0064\% MAE, respectively. In other words, the higher the approximation error is, the higher is the inexact nature of DNNs and hence, the lower is their classification accuracy and fault resilience. In this paper, the KR3-based AxDNN is not analyzed due to its quite high MAE (i.e., 3.08\%) and low classification accuracy (see Fig. \ref{fig:maeAndAcc} (a) and Fig. \ref{fig:maeAndAcc} (b))

\begin{figure*}[!b]
	\centering
	\vspace{-0.01in}
	\includegraphics[width=1\linewidth]{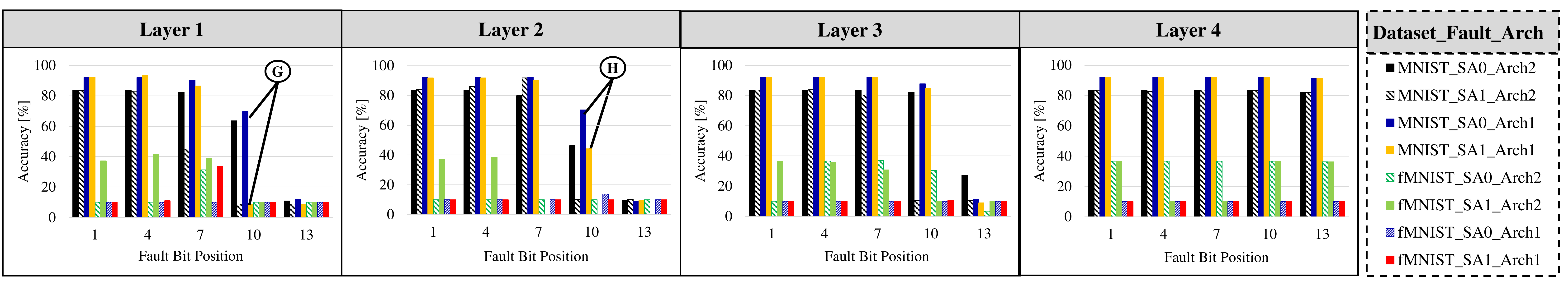}
	\caption{Stuck-at Faults Resilience Analysis of L2D-based Approximate DNNs (AxDNN) \cite{mrazek2017evoapproxsb}, using the MNIST \cite{deng2012mnist} and Fashion MNIST \cite{xiao2017fashion} datasets.}
	\label{fig:L2D}
	\vspace{-0.1in}
\end{figure*}

\begin{figure*}[!b]
	\centering
	\vspace{-0.01in}
	\includegraphics[width=1\linewidth]{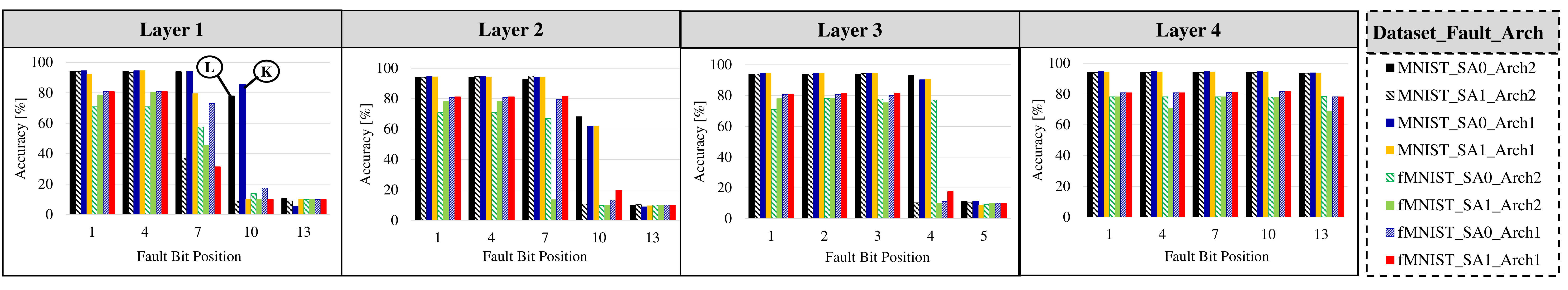}
	\caption{Stuck-at Faults Resilience Analysis of KX2-based Approximate DNNs (AxDNN) \cite{mrazek2017evoapproxsb}, using the MNIST \cite{deng2012mnist} and Fashion MNIST \cite{xiao2017fashion} datasets.}
	\label{fig:KX2}
	\vspace{-0.1in}
\end{figure*}

\begin{figure*}[!b]
	\centering
	\vspace{-0.01in}
	\includegraphics[width=1\linewidth]{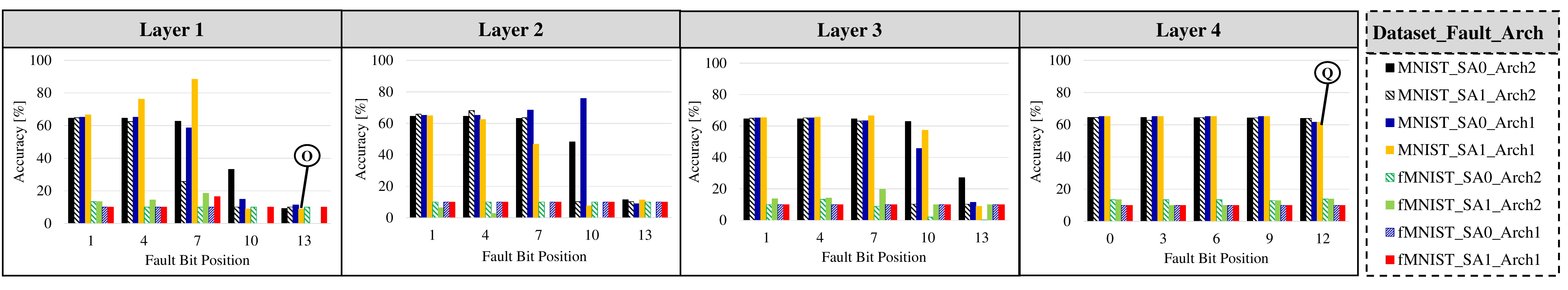}
	\caption{Stuck-at Faults Resilience Analysis of L1G-based Approximate DNNs (AxDNN) \cite{mrazek2017evoapproxsb}, using the MNIST \cite{deng2012mnist} and Fashion MNIST \cite{xiao2017fashion} datasets.}
	\label{fig:L1G}
	\vspace{-0.15in}
\end{figure*}

\begin{figure*}[!b]
	\centering
	\vspace{-0.01in}
	\includegraphics[width=1\linewidth]{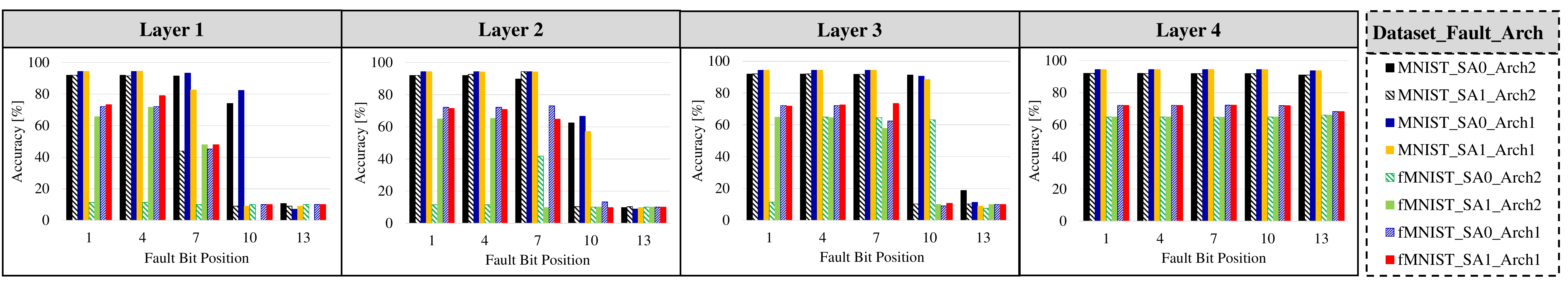}
	\caption{Stuck-at Faults Resilience Analysis of KRC-based Approximate DNNs (AxDNN) \cite{mrazek2017evoapproxsb}, using the MNIST \cite{deng2012mnist} and Fashion MNIST \cite{xiao2017fashion} datasets.}
	\label{fig:KRC}
	\vspace{-0.1in}
\end{figure*}

\begin{figure*}[!b]
	\centering
	\vspace{-0.01in}
	\includegraphics[width=1\linewidth]{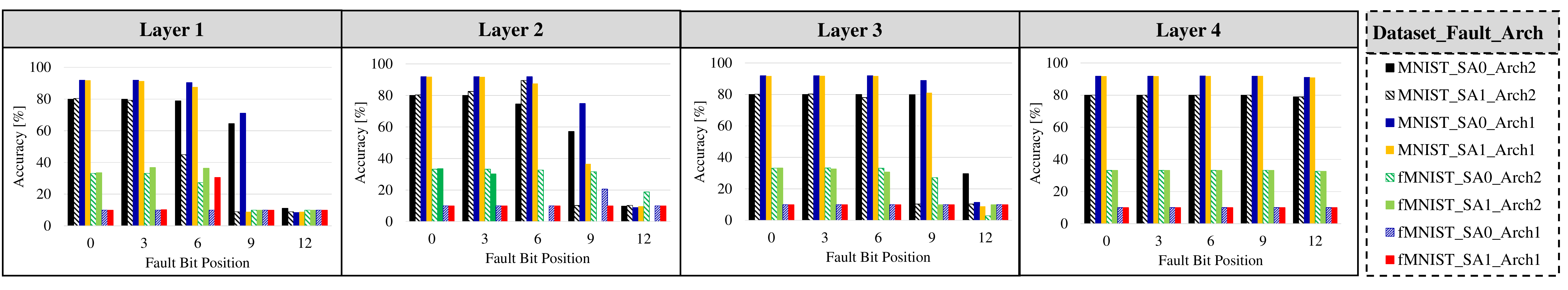}
	\caption{Stuck-at Faults Resilience Analysis of KVL-based Approximate DNNs (AxDNN) \cite{mrazek2017evoapproxsb}, using the MNIST \cite{deng2012mnist} and Fashion MNIST \cite{xiao2017fashion} datasets.}
	\label{fig:KVL}
	\vspace{-0.1in}
\end{figure*}

\begin{figure*}[!b]
	\centering
	\vspace{-0.01in}
	\includegraphics[width=1\linewidth]{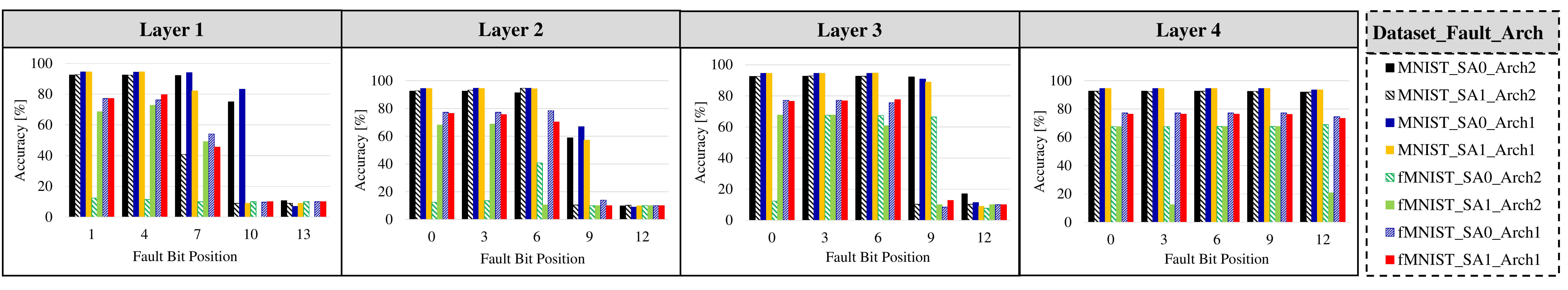}
	\caption{Stuck-at Faults Resilience Analysis of KR6-based Approximate DNNs (AxDNN) \cite{mrazek2017evoapproxsb}, using the MNIST \cite{deng2012mnist} and Fashion MNIST \cite{xiao2017fashion} datasets.}
	\label{fig:KR6}
	\vspace{-0.15in}
\end{figure*}

\begin{figure*}[!t]
	\centering
	\vspace{-0.15in}
	\includegraphics[width=1\linewidth]{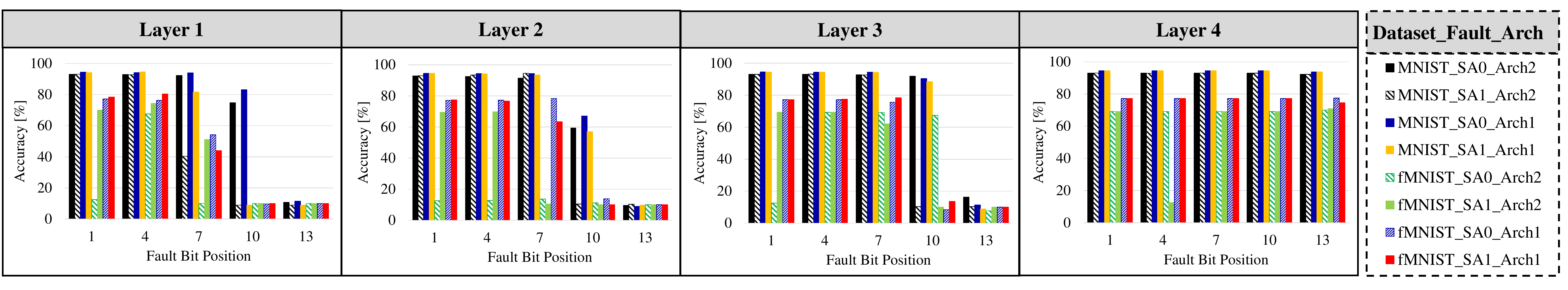}
	\caption{Stuck-at Faults Resilience Analysis of L2H-based Approximate DNNs (AxDNN) \cite{mrazek2017evoapproxsb}, using the MNIST \cite{deng2012mnist} and Fashion MNIST \cite{xiao2017fashion} datasets.}
	\label{fig:L2H}
	\vspace{-0.2in}
\end{figure*}

\subsubsection{Fault Configuration vs Fault Resilience} 
\label{subsubsec:dconfiganal}
Similar to the accurate DNNs, the type and location of the faults may affect the output of the AxDNNs. The faulty MSBs in both accurate DNNs and AxDNNs may lead to substantial accuracy loss (e.g., approximately 98\% to 100\%), by changing the desired values significantly, as compared to the faulty LSBs. The MNIST \cite{deng2012mnist} and Fashion-MNIST \cite{xiao2017fashion} classification with a fault in the LSB of accurate multipliers in accurate DNNs may lead to approximately 0\% to 1\% accuracy loss in comparison to the non-faulty accurate DNNs (see Fig \ref{fig:maeAndAcc} (b) and Fig. \ref{fig:KV8}). However, it may lead to approximately 0\% to 9\% accuracy loss in AxDNNs in comparison to non-faulty AxDNNs (as shown in Fig. \ref{fig:maeAndAcc} (b) and Fig. \ref{fig:KVA_v2} to Fig. \ref{fig:L2H}). Interestingly, the MNIST \cite{deng2012mnist} classification with stuck-at-0 fault in the tenth bit of the accurate multiplier-based Arch. 1 (fault in layer 1 only) results in 86.54\% accuracy (see label E in Fig. \ref{fig:KV8}) but the same fault configuration in KTY-based AxDNN leads to 16.3\% accuracy only (see label F in Fig. \ref{fig:KTY}). Hence, the faults exacerbates the accuracy loss with approximate computing in AxDNNs and the fault resilience decreases from LSB to MSB. The same trend is observed in case of Fashion-MNIST dataset. Furthermore, it is observed that stuck-at-0 faults contribute comparatively less towards the output quality degradation as compared to the stuck-at-1 faults, on average, in both accurate DNNs and AxDNNs. This effect is more noticeable in DNN layers 1 and 2 (e.g., see label G and H in Fig. \ref{fig:L2D}). 

\subsubsection{DNN Design Configuration vs Fault Resilience}
\label{subsubsec:fconfiganal}
Similar to the accurate DNNs, the impact of faults may vary with the type of activation functions used in AxDNNs. Our quantitative fault resilience analysis shows that Arch. 1 with \textit{tanh} activation function, in its hidden layers, seems to be comparatively less disturbed by the faults, on average, as compared to Arch. 2 with \textit{sigmoid} activation function. For example, the MNIST \cite{deng2012mnist} classification with stuck-at-0 fault in layer 1 (and tenth bit) of the accurate multiplier-based Arch. 1 and Arch. 2 results in 8.27\% (see label A in Fig. \ref{fig:maeAndAcc} (b) and label E in Fig. \ref{fig:KV8}) and 15.14\% (see label I in Fig. \ref{fig:maeAndAcc} (b) and label J in Fig. \ref{fig:KV8}) accuracy loss, respectively, as compared to their corresponding non-faulty accurate counterparts. Likewise, the same fault configuration with KX2-based Arch. 1 and Arch. 2 yields 9.12\% (see label A in Fig. \ref{fig:maeAndAcc} (b) and label K in Fig. \ref{fig:KX2}) and 16.41\% (see label I in Fig. \ref{fig:maeAndAcc} (b) and label L in Fig. \ref{fig:KX2}) accuracy loss, respectively. In comparison to the non-faulty approximate counterparts (see label M and N in Fig. \ref{fig:maeAndAcc} (b)), these faults lead to 8.84\% and 16\% accuracy loss in Arch. 1 and Arch. 2, respectively, in the presence of approximation error. The same trend is observed in case of Fashion-MNIST dataset. Hence, the faults trend remains the same for different DNN architectures regardless of their accurate and approximate nature. Moreover, it is observed that the input layer is quite less resilient to the faults as compared to the output layer. This effect is more visible in MSBs (e.g., see layer 1 and 4 in Fig. \ref{fig:KVB}). The reason is that faults in the input layer may affect the output of all DNN layers and decrease the fault resilience significantly. For example, the MNIST classification with a stuck-at-1 fault in thirteenth bit of the L1G approximate multiplier, located in layer 1, of Arch. 1 can lead to approximately 82\% and 56\% accuracy loss, respectively, in comparison to non-faulty accurate DNN (see label A in Fig. \ref{fig:maeAndAcc} (b) and label O in Fig. \ref{fig:L1G}) and AxDNN (see label P in \ref{fig:maeAndAcc} (b) and label O in Fig. \ref{fig:L1G}), respectively. However, the same fault configuration in layer 4 can lead to approximately 30\% and 3\% accuracy loss only in comparison to non-faulty accurate DNN (see label A in Fig. \ref{fig:maeAndAcc} (b) and label Q in Fig. \ref{fig:L1G}) and AxDNN (see label O in Fig. \ref{fig:maeAndAcc} (b) and label Q in \ref{fig:L1G}). Likewise, the Fashion-MNIST \cite{xiao2017fashion} classification with similar fault configuration in layer 1 and 4 of KTY-based Arch. 2 leads to approximately 80\% (see label R in Fig. \ref{fig:KTY} and label S in Fig. \ref{fig:maeAndAcc} (b)) and 66\% (see label T in Fig. \ref{fig:KTY} and label S in Fig. \ref{fig:maeAndAcc} (b)) accuracy loss in comparison to non-faulty accurate DNNs, respectively. Conversely, it has 9\% accuracy loss only in faulty accurate DNN (see label S in Fig. \ref{fig:maeAndAcc} (b) and label U in Fig. \ref{fig:KV8}). 


 

\subsection{Fault-Energy Trade-off Exploration}
\begin{figure}[!h]
	\centering
	\vspace{-0.15in}
	\includegraphics[width=0.5\linewidth]{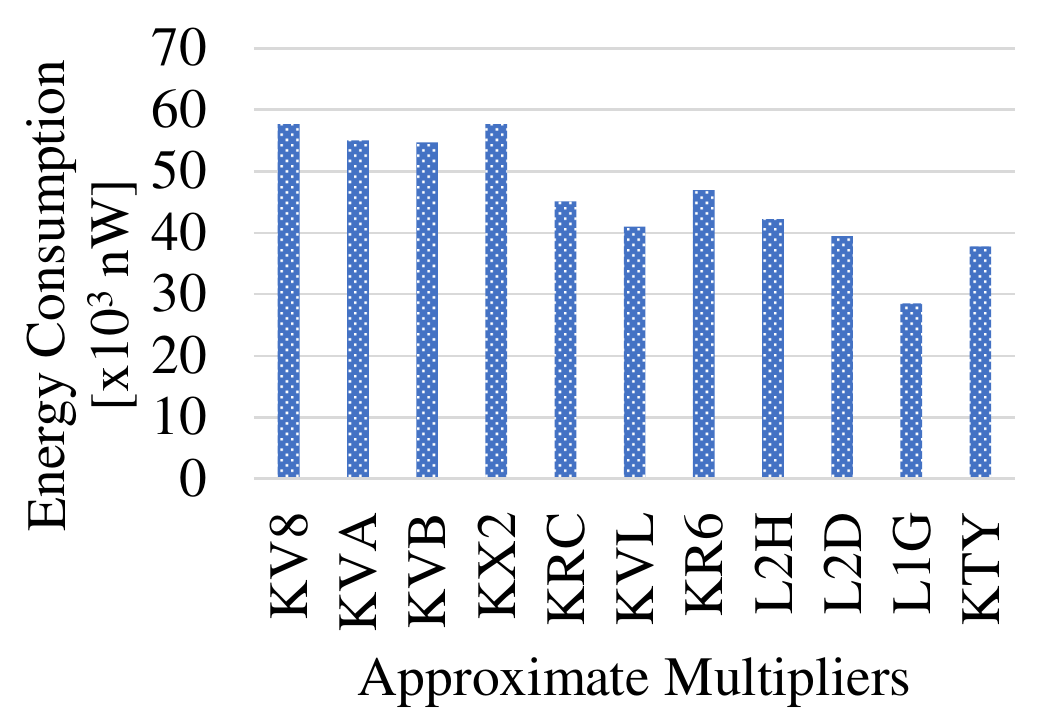}
	\vspace{-3mm}
	\caption{Energy Analysis of Approximate 8x8 Systolic Arrays using Evoapprox8b \cite{mrazek2017evoapproxsb} Signed Multipliers} 
	\label{fig:energyAnalysis}
	\vspace{-0.15in}
\end{figure}

Fig.\ref{fig:energyAnalysis} compares the energy efficiency of the Evoapprox8b \cite{mrazek2017evoapproxsb} signed multipliers-based 8x8 systolic arrays \cite{zhang2018analyzing}. It is evident from this analysis that approximate computing increases the energy efficiency of DNNs. The Evoapprox8b \cite{mrazek2017evoapproxsb} approximate multipliers are designed in such a way that they provide low power or latency or both. Our fault resilience and energy analysis reveals that the KVA- and KVB-based DNNs are the most fault resilient but less energy efficient among the Evoapprox8b-based AxDNNs. Furthermore, the L2D, L1G- and KTY-based AxDNNs are least fault resilient but most energy efficient due to their comparatively low classification accuracy and energy consumption. \textit{Hence, the fault resilience and energy efficiency are orthogonal to each other.} As discussed earlier, the fault resilience decreases significantly with faults in the input layer and MSB (as discussed in Section \ref{faultRes}). Hence, the KRC-based AxDNNs seems to be quite energy efficient and fault resilient as it exhibit above 90\% and 80\% accuracy, with faulty LSB in the input layer, in the MNIST \cite{deng2012mnist} and Fashion MNIST \cite{xiao2017fashion} classification (as shown in Fig. \ref{fig:KRC}), respectively. 

\section{Conclusion}
\label{sec:conclusion}
Approximate computing relaxes the abstraction of near-perfect accuracy in error resilient applications to improve their energy efficiency. However, such inexactness may reduce the fault resilience in DNNs. Towards this, we explore fault-energy trade-offs in approximate FFNNs, using the state-of-the-art Evoapprox8b \cite{mrazek2017evoapproxsb} signed multipliers, by varying the stuck-at-0 and stuck-at-1 fault-bit positions and using different activation functions (e.g., \textit{tanh} and \textit{sigmoid}) in the hidden layers. Our results demonstrate that the faults exacerbates the accuracy loss, from LSB to MSB and output to input layer, with approximate computing in AxDNNs. Their impact also varies with the activation functions. We observe that analysis reveals that the fault resilience and energy efficiency of AxDNNs are orthogonal to each other.

\balance

\bibliographystyle{IEEEtran}
\bibliography{bib/conf}

\end{document}